\documentclass[11pt,a4paper]{article}
\usepackage[hyperref]{acl2021}
\usepackage{times}
\usepackage{latexsym}
\usepackage{soul}
\usepackage{url}
\usepackage{amsmath}
\usepackage{amsthm}
\usepackage{amssymb}
\usepackage{multirow}
\usepackage{bm}
\usepackage{color, xcolor}
\usepackage{booktabs}
\usepackage{algorithm}
\usepackage{algorithmic}
\usepackage{graphicx}

\usepackage{microtype}

\aclfinalcopy 

\title{Multi-hop Graph Convolutional Network with High-order Chebyshev Approximation for Text Reasoning}

\author{Shuoran Jiang, \ \ Qingcai Chen\thanks{\ \ corresponding author: Qingcai Chen}, \ \ Xin Liu, \ \ Baotian Hu, \ \ Lisai Zhang \\ Harbin Institute of Technology, Shenzhen \\ \texttt{\{shuoran.chiang, hit.xinliu, lisaizhang2016\}@gmail.com}\\ \texttt{\{qingcai.chen, hubaotian\}@hit.edu.cn}}

\date{}

\begin{document}
\maketitle
\begin{abstract}
Graph convolutional network (GCN) has become popular in various natural language processing (NLP) tasks with its superiority in long-term and non-consecutive word interactions.
However,
existing single-hop graph reasoning in GCN may miss some important non-consecutive dependencies.
In this study,
we define the spectral graph convolutional network with the high-order dynamic Chebyshev approximation (HDGCN),
which augments the multi-hop graph reasoning by fusing messages aggregated from direct and long-term dependencies into one convolutional layer.
To alleviate the over-smoothing in high-order Chebyshev approximation,
a multi-vote based cross-attention (MVCAttn) with linear computation complexity is also proposed.
The empirical results on four \emph{transductive} and \emph{inductive} NLP tasks and the ablation study verify the efficacy of the proposed model.
Our source code is available at \url{https://github.com/MathIsAll/HDGCN-pytorch}.
\end{abstract}

\section{Introduction}

Graph neural networks (GNNs) are usually used to learn the node representations in Euclidean space from graph data,
which have been developed to one of the hottest research topics in recent years ~\cite{zhang2020get}.
The primitive GNNs relied on recursive propagation on graphs,
which takes a long time to train ~\cite{zhang2019bayesian}.
One major variant of GNNs, graph convolutional networks (GCNs) ~\cite{kipf2016semi,yao2019graph},
takes spectral filtering to replace recursive message passing and needs only a shallow network to convergent,
which have been used in various NLP tasks.
For example,
~\citet{yao2019graph}
constructed the text as a graph and input it to a GCN.
This method achieved better results than conventional deep learning models in text classification.
Afterward,
the GCNs have became popular in more tasks,
such as word embedding~\cite{zhang2020every},
semantic analysis~\cite{zhang2019aspect},
document summarization~\cite{wang2020heterogeneous},
knowledge graph~\cite{wang2018cross},
etc.

\begin{figure}[h]
	\centering
	\includegraphics[scale=0.045]{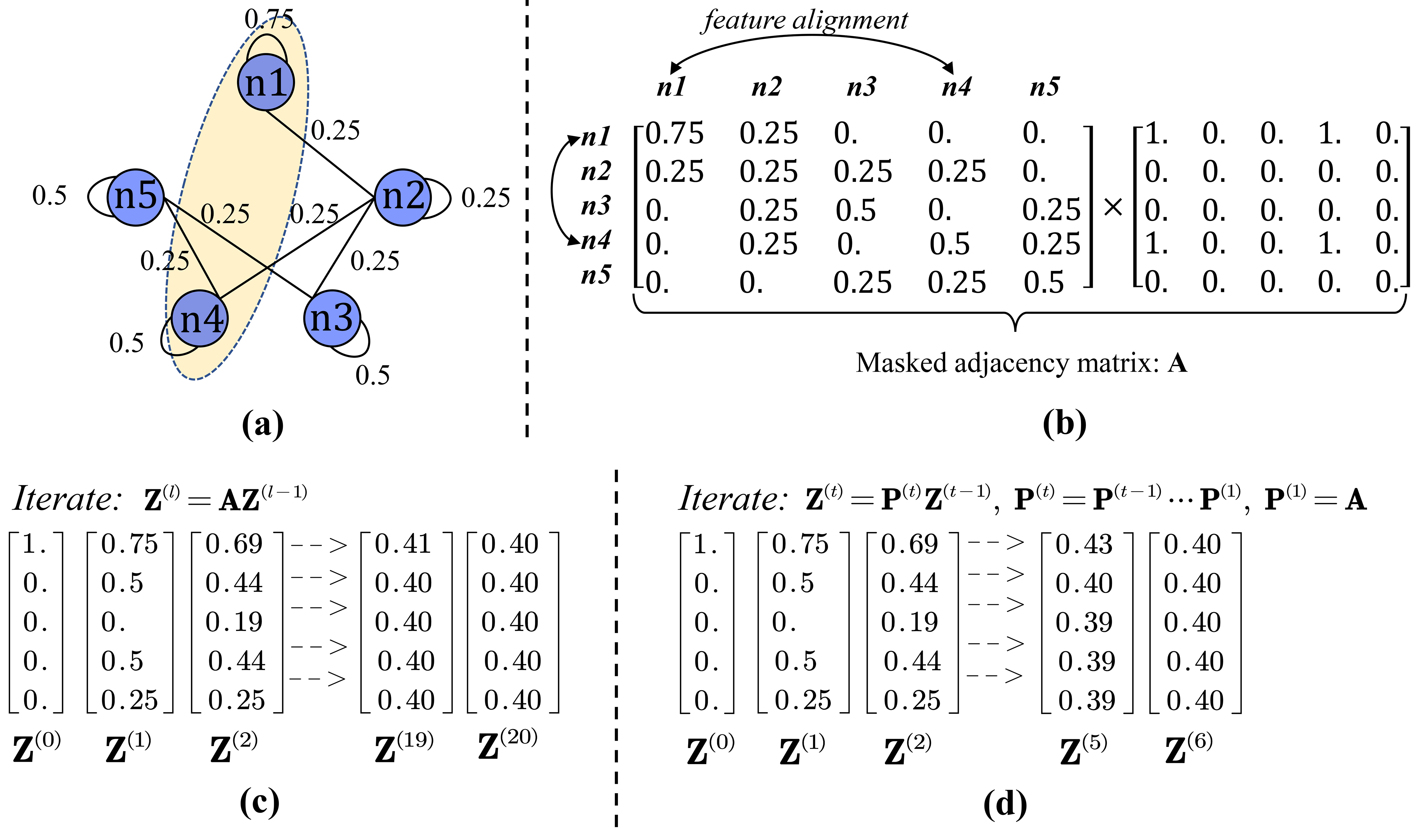}
	\caption{\textbf{(a):} A simple graph with 5 nodes and the weighted edges, in which the nodes $n4$ is a pronoun of $n1$ and the two nodes need to align features. \textbf{(b):} The masked adjacency matrix on this graph. \textbf{(c)} and \textbf{(d):} The processes of feature alignment on nodes $n1$ and $n4$ without transition probability and with transition probability respectively.}
	\label{mhr_demo}
\end{figure}

The spectral graph convolution in Yao's GCN is a localized first-order Chebyshev approximation.
It is equal to a stack of 1-step Markov chain (MC) layer and fully connected (FC) layer.
Unlike the multi-step Markov chains,
the message propagation in vanilla GCN lacks the node probability transitions.
As a result,
the multi-hop graph reasoning is very tardy in GCN and easily causes the \emph{suspended animation problem}~\cite{zhang2019gresnet}.
However,
the probability transition on the graph is useful to improve the efficiency in learning contextual dependencies.
In many NLP tasks (like the question answering (QA) system and entity relation extraction),
the features of the two nodes need to be aligned.
As an example,
Figure \ref{mhr_demo} shows a simple graph where the node $n4$ is a pronoun of node $n1$.
In this example,
the adjacency matrix is masked on nodes $n2$, $n3$, $n5$ to demonstrate the message passing between $n1$ and $n4$.
Figure \ref{mhr_demo} (c) and (d) plot the processes of feature alignment on two nodes without and with probability transitions respectively.
In this example,
the feature alignment process without probability transition needs 10 more steps than which with probability transition.
It is shown that encoding the multi-hop dependencies through the spectral graph filtering in GCN usually requires a deep network.
However,
as well known that the deep neural network (DNN) is tough to train and easily causes the over-fitting problem~\cite{rong2019dropedge}.

Some newest studies to improve the \emph{multi}-\emph{hop graph reasoning} include graph attention networks (GATs) ~\cite{velivckovic2018graph},
graph residual neural network (GRESNET) ~\cite{zhang2019gresnet},
graph diffusive neural network (DIFNET) ~\cite{zhang2020get}, TGMC-S ~\cite{zhang2020network} and Graph Transformer Networks ~\cite{yun2019graph, zhang-zhang-2020-text}.
GATs enhance the graph reasoning by implicitly re-defining the graph structure with the attention on the 1-hop neighbors,
but there is equilibrial optimization on the whole graph.
GRESNET solves the \emph{suspended animation problem} by creating extensively connected highways to involve raw node features and intermediate representations throughout all the model layers.
However,
the multi-hop dependencies are still reasoned at a slow pace.
DIFNET introduces a new neuron unit,
i.e.,
GDU (gated diffusive unit),
to model and update the hidden node states at each layer.
DIFNET replaces the spectral filtering with a recursive module and realizes the neural gate learning and graph residual learning.
But the time cost is aggravated in DIFNET compared with GCN.
TGMC-S stacks GCN layers on adjacent matrices with different hops of traffic networks.
Different from the ground-truth traffic network in TGMC-S,
it is hard to construct the multi-hop word-word relationships objectively from the text.
TGMC-S hadn’t given a way to improve the multi-hop message passing in GCN.

Transformers~\cite{vaswani2017attention} and corresponding pre-trained models~\cite{xu2019bert} could be thought of as fully-connected graph neural networks that contain the multi-hop dependencies.
They figure out the contextual dependencies on the fully-connected graph with the attention mechanism.
The message propagation in transformers follows the relations self-adaptively learned from input sequence instead of the fixed graph structures.
Publications have shown that transformers outperform GCNs in many NLP tasks.
Graph Transformer ~\cite{dwivedi2020generalization} generalizes the Transformer to arbitrary graphs,
and improves \emph{inductive learning} from Laplacian eigenvectors on graph topology.
However,
due to the connections scale quadratically growth with node number $N$ in graphs,
things get out of hand for very large $N$.
Additionally,
the fully-connected graph is not an interpretable architecture in practical tasks.
For example,
whether Transformers are the best choice to bring the text in linguistic theory? \footnote{https://towardsdatascience.com/transformers-are-graph-neural-networks-bca9f75412aa}

To improve the efficiency and performance of \emph{multi}-\emph{hop graph reasoning} in spectral graph convolution,
we proposed a new graph convolutional network with high-order dynamic Chebyshev approximation (HDGCN).
A prime ChebNet and a high-order dynamic (HD) ChebNet are firstly applied to implement this Chebyshev approximation.
These two sub-networks work like a trade-off on low-pass signals (direct dependencies) and high-pass signals (multi-hop dependencies) respectively.
The prime ChebNet takes the same frame as the convolutional layer in vanilla GCN.
It mainly extracts information from direct neighbors in local contexts.
The HD-ChebNet aggregates messages from multi-hop neighbors following the transition direction adaptively learned by the attention mechanism.
The standard self-attention \cite{vaswani2017attention} has a $\mathcal{O}\left( N^2 \right)$ computation complexity and it is hard to be applied on long sequence.
Even the existing sparsity attention methods,
like the Star-Transformer \cite{guo2019star} and Extended Transformer Construction (ETC) \cite{ainslie2020etc},
have reduced the quadratic dependence limit of sequence length to linear dependence,
but the fully-connected graph structure cannot be kept.
We design a multi-vote-based cross-attention (MVCAttn) mechanism.
The MVCAttn scales the computation complexity $\mathcal{O}(N^2)$ in self-attention to $\mathcal{O}(N)$.

The main contributions of this paper are listed below:
\begin{itemize}
	\item To improve the efficiency and performance of multi-hop reasoning in spectral graph convolution, we propose a novel graph convolutional network with high-order dynamic Chebyshev Approximation (HDGCN).
	\item To avoid the over-smoothing problem in HD-ChebNet, we propose a multi-vote based cross-attention (MVCAttn) mechanism, which adaptively learn the direction of node probability transition. MVCAttn is a variant of the attention mechanism with the property of linear computation complexity.
	\item The experimental results show that the proposed model outperforms compared SOTA models on four transductive and inductive NLP tasks.
\end{itemize}

\section{Related Work}

Our work draws supports from the vanilla GCN and the attention mechanism,
so we first give a glance at the paradigm of these models in this section.

\subsection{Graph Convolutional Network}

The GCN model proposed by ~\cite{kipf2016semi} is the one we interested,
and it is defined on graph $\mathcal{G}=\lbrace \mathcal{V}, \mathcal{E} \rbrace$,
where $\mathcal{V}$ is the node set and $\mathcal{E}$ is the edge set.
The edge $\left( v_i, v_j \right) \in \mathcal{E}$ represents a link between nodes $v_i$ and $v_j$.
The graph signals are attributed as $\mathbf{X} \in \mathbb{R}^{\vert \mathcal{V} \vert \times d}$,
and the graph relations $\mathcal{E}$ can be defined as an adjacency matrix $\mathbf{A} \in \mathbb{R}^{\vert \mathcal{V} \vert \times \vert \mathcal{V} \vert}$ (binary or weighted).

Each convolutional layer in GCN is a 1st Chebyshev approximation on spectral graph convolution,
and its layer-wise propagation rule in neural network is defined as:
\begin{equation}
{\small
	\begin{aligned}
		\mathbf{H}^{(l+1)} &= \sigma \left( \widetilde{\mathbf{A}} \mathbf{H}^{(l)} \mathbf{W}^{(l)} \right), \quad L \geq l \geq 0 \\
		\widetilde{\mathbf{A}} &= \left( \mathbf{D} + \mathbf{I}_N \right)^{-\frac{1}{2}} \left( \mathbf{A} + \mathbf{I}_N \right) \left( \mathbf{D} + \mathbf{I}_N \right)^{-\frac{1}{2}},
	\end{aligned}
	\label{layer_wise_gcn}}
\end{equation}
where $\mathbf{H}^{(0)}=\mathbf{X}$,
$\widetilde{\mathbf{A}}$ is the normalized adjacency matrix and $\sigma$ is a non-linear activation function.

The node embeddings output from the last convolutional layer are fed into a \emph{softmax} classifier for node or graph classification,
and the loss function $\mathcal{L}$ can be defined as the cross-entropy error.
The weight set $\lbrace \mathbf{W}^{(l)} \rbrace^L_{l=0}$ can be jointly optimized by minimizing $\mathcal{L}$ via gradient descent.

\subsection{Self-Attention Is a Dynamic GCN}

The attention mechanism is an effective way to extract task-relevant features from inputs,
and it helps the model to make better decisions ~\cite{lee2019attention}.
It has various approaches to compute the attention score from features,
and the scaled dot-product attention proposed in Transformers ~\cite{vaswani2017attention} is the most popular one.
\begin{equation}
{\small
	\begin{aligned}
		\mathbf{Z} = \underset{\mathbf{A}}{\underbrace{\mbox{softmax} \left( \frac{\mathbf{XW}^q\mathbf{W}^k\mathbf{X}^{\mathrm{T}}}{\sqrt{d_k}} \right)}} \mathbf{X} \mathbf{W}^v
	\end{aligned}
	\label{self_attn_gcn}}
\end{equation}
where $\mathbf{X} \in \mathbb{R}^{N \times d}$ is the input sequence,
and weights $\mathbf{W}^q \in \mathbb{R}^{d \times d_k}$, $\mathbf{W}^k \in \mathbb{R}^{d_k \times d}$, $\mathbf{W}^v \in \mathbb{R}^{d \times d_v}$ are used to transform sequence to queries, keys and values.

As showed in Equation \ref{self_attn_gcn},
the attention scores $\mathbf{A}$ can be viewed as a dynamic adjacency matrix on sequence $\mathbf{X}$.
This process in self-attention is similar to the graph convolutional layer defined in Equation \ref{layer_wise_gcn}.
The only difference is that the adjacency matrix in Equation \ref{self_attn_gcn} is adaptively learned from input instead of prior graph structures.

\section{Method}

\begin{figure*}[h]
	\centering
	\includegraphics[scale=0.065]{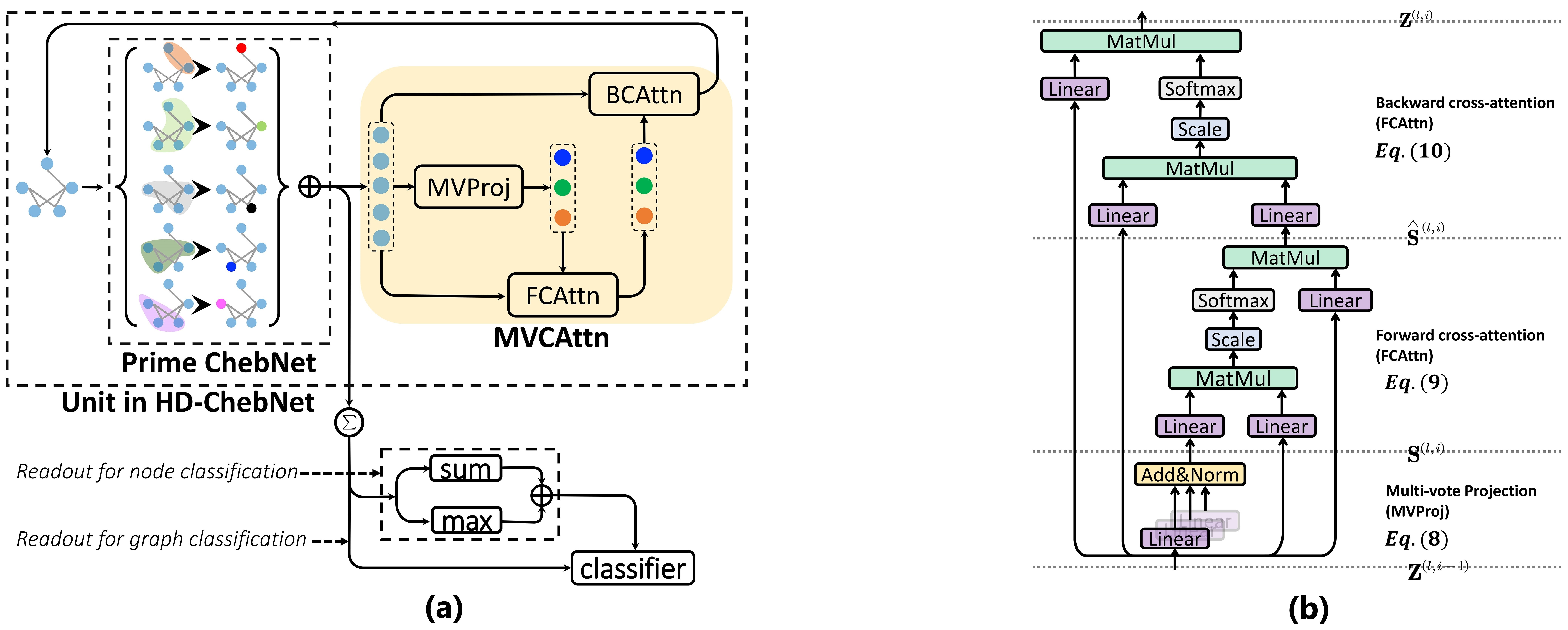}
	\caption{\textbf{(a):} The architecture of HDGCN taking the simple graph in Figure \ref{mhr_demo} as an example. \textbf{(b):} The schematics of the multi-vote based cross-attention (MVCAttn) in every unit in HD-ChebNet.}
	\label{mvcattn_schem}
\end{figure*}

In our model,
the input graph $\mathcal{G}= \left( \mathcal{V}, \mathcal{E} \right)$ takes the same form as the one in GCN.
The nodes are attributed as $\mathbf{X} \in \mathbb{R}^{\vert \mathcal{V} \vert \times d}$,
and the adjacency matrix $\mathbf{A} \in \mathbb{R}^{\vert \mathcal{V} \vert \times \vert \mathcal{V} \vert}$ (binary or weighted) is defined on graph edges $\mathcal{E}$.

The spectral graph convolution in Fourier domain is defined as,
\begin{equation}
{\small
	\bm{g}_{\boldsymbol{\theta}} \star \bm{x} = \mathbf{U} \bm{g}_{\boldsymbol{\theta}} \left(\widetilde{\boldsymbol{\Lambda}} \right) \mathbf{U}^T \bm{x}
	\label{graphconv}}
\end{equation}
where $\bm{x} \in \mathbb{R}^d$ is the signal on a node,
$\mathbf{U}$ is the matrix of eigenvectors on normalized graph Laplacian $\mathbf{L}=\mathbf{I}_N - \mathbf{D}^{-\frac{1}{2}} \mathbf{A} \mathbf{D}^{-\frac{1}{2}} = \mathbf{U} \boldsymbol{\Lambda} \mathbf{U}^T$,
and the filter $\bm{g}_{\boldsymbol{\theta}}(\widetilde{\boldsymbol{\Lambda}})$ is a function of the eigenvalues on normalized $\widetilde{\mathbf{L}}$ in Fourier domain.

The $K$-th ($K>2$) order truncation of Chebyshev polynomials on this spectral graph convolution is,
\begin{equation}
{\small
	\bm{g}_{\boldsymbol{\theta}} \star \bm{x} \approx \sum_{i=0}^K \boldsymbol{\theta}_i \mathbf{U} \bm{T}_i \left(\widetilde{\boldsymbol{\Lambda}} \right) \mathbf{U}^T \bm{x}
	\label{korder}}
\end{equation}
where {\small $\bm{T}_0\left( \widetilde{\mathbf{\Lambda }} \right) =\mathbf{I}$},
{\small $\bm{T}_1=\widetilde{\mathbf{\Lambda }}$},
{\small $\bm{T}_{i>1}\left( \widetilde{\mathbf{\Lambda }} \right) =2\widetilde{\mathbf{\Lambda }}\bm{T}_{i-1}\left( \widetilde{\mathbf{\Lambda }} \right) - \bm{T}_{i-2}\left( \widetilde{\mathbf{\Lambda }} \right)$}.

To replace the parameters $\lbrace \boldsymbol{\theta}_i \rbrace^K_{i=1}$ with another parameter set $\lbrace \boldsymbol{\theta}^{(i)} \rbrace^{K/2}_{i=1}$,
the $K$th-order Chebyshev polynomials in Equation \ref{korder} are approximated as:
\begin{equation}
{\small
	\begin{aligned}
		\bm{g}_{\boldsymbol{\theta}} \star \bm{x} &\approx \sum^{K/2}_{k=0} \left( \mathbf{U} \widetilde{\boldsymbol{\Lambda}} \mathbf{U}^T \right)^{2k} \left( \mathbf{I} - \mathbf{U} \widetilde{\boldsymbol{\Lambda}} \mathbf{U}^T \right) \bm{x} \boldsymbol{\theta}^{(k)} \\
		&\approx \sum^{K/2}_{k=1} \widetilde{\mathbf{A}}^{2k} \widetilde{\mathbf{A}} \bm{x} \boldsymbol{\theta}^{(i)}
	\end{aligned}}
\label{3rdcheb}
\end{equation}
where the $\widetilde{\mathbf{A}}$ is normalized adjacency matrix (as defined in Equation \ref{layer_wise_gcn}).
As the node state transition $\widetilde{\mathbf{A}}^{2k}$ causes the over-smoothing problem \cite{li2018deeper, nt2019revisiting},
we take the dynamic pairwise relationship $\mathbf{A}_d$ self-adaptively learned by the attention mechanism to turn the direction of node state transition.

The powers of adjacency matrix $\widetilde{\mathbf{A}}^{2k}$ in Equation \ref{3rdcheb} can cause the over smoothing problem,
we replace the $\widetilde{\mathbf{A}}^{2k}$ with $\widetilde{\mathbf{A}}^k \mathbf{A}^k_d$.

In our implementation,
the first-order and higher-order Chebyshev polynomials in Equation \ref{3rdcheb} is approximated with a prime Chebyshev network (ChebNet) and high-order dynamic Chebyshev networks (HD-ChebNets) respectively.
We generalize the graph convolution on $K$th-order dynamic Chebyshev approximation (Equation \ref{3rdcheb}) to the layer-wise propagation as follows,
\begin{equation}
{\small
	\begin{aligned}
		\mathbf{H} &\approx \sum^{K/2}_{k=0} \mathbf{Z}^{(k)}, \\
		\mathbf{Z}^{(0)} &= \underset{\mbox{Prime ChebNet}}{\underbrace{\sigma \left( \widetilde{\mathbf{A}} \mathbf{X} \mathbf{W}^{(0)} \right)}}, \\
		\mathbf{Z}^{(k)} &= \underset{\mbox{Unit in HD-ChebNet}}{\underbrace{\sigma \left( \widetilde{\mathbf{A}} \left( \mathbf{A}_{d}^{(k)} \mathbf{Z}^{(k)}\mathbf{W}^{(k)}_d \right) \mathbf{W}^{(k)} \right)}},
	\end{aligned}}
	\label{final_define}
\end{equation}
where $k$ is the order and $\mathbf{W}^{(0)}$, $\mathbf{W}^{(k)}$, $\mathbf{W}^{(k)}_d$ are nonlinear filters on node signals.
For the convenience of writing,
we just define the first layer of HDGCN.

\subsection{Prime ChebNet}

We consider the same convolutional architecture as the one in GCN to implement the prime ChebNet,
and it mainly aggregates messages from the direct dependencies.
\begin{equation}
{\small 
	\begin{aligned}
		\mathbf{Z}^{(0)} &= \sigma \left( \widetilde{\mathbf{A}} \mathbf{X} \mathbf{W}^{(0)} \right),
	\end{aligned}}
	\label{gcns}
\end{equation}
where $\mathbf{W}^{(0)} \in \mathbb{R}^{d \times d}$ and $\widetilde{\mathbf{A}}$ is the normalized symmetric adjacency matrix.

\subsection{High-order Dynamic (HD) ChebNet}

As the multi-hop neighbors can be interacted via the 1-hop neighbors,
we take the $\mathbf{Z}^{(0)}$ output from the prime ChebNet as input of the HD-ChebNet.
The multi-vote based cross-attention (MVCAttn) mechanism first adaptively learns the direction of node probability transition $\mathbf{A}^{(k)}_d$,
its schematic is showed in Figure \ref{mvcattn_schem} (b).
MVCAttn has two phases - graph information aggregation and diffusion.

\textbf{\emph{Graph Information Aggregation}} coarsens the node embeddings $\mathbf{Z}^{(k-1)}$ to a small supernode set $\mathbf{S}^{(k)} \in \mathbb{R}^{M \times d}$,
$M \ll \vert \mathcal{V} \vert$.

The first step is multi-vote projection (MVProj).
In which node embeddings $\mathbf{Z}^{(k-1)}$ are projected to multiple votes $\mathbf{V}^{(k)} \in \mathbb{R}^{\vert \mathcal{V} \vert \times M \times d}$,
and these votes are aggregated to supernode set $\mathbf{S}^{(k)} = \lbrace \bm{s}^{(k)}_{m} \rbrace_{m=1}^{M}$.
\begin{equation}
	\begin{aligned}
		\bm{s}^{(k)}_m &= \mbox{MVProj}\left( \mathbf{Z}^{(k-1)} \right) \\
		& = \mbox{norm} \left( \sum^{\vert \mathcal{V} \vert}_{v=1} \bm{z}^{(k-1)}_v \mathbf{W}_m^V \right)
	\end{aligned}
\end{equation}
where $\vert \mathcal{V} \vert \geq v \geq 1$, $M \geq m \geq 1$,
$\mathbf{W}_m^V \in \mathbb{R}^{d_k \times d_k}$ is the projection weight and $\mbox{norm} \left( \right)$ represents the \textbf{LayerNorm} operation.

Next,
the forward cross-attention (FCAttn) updates the supernode values as:
\begin{equation}
	\begin{aligned}
		\widehat{\mathbf{S}}^{(k)} &= \mbox{FCAttn} \left( \mathbf{Z}^{(k)}, \mathbf{S}^{(k)} \right) \\
		&= \mathbf{A}^{(k)}_f \mathbf{Z}^{(k-1)} \mathbf{W}_{fv} \\
		\mathbf{A}^{(k)}_f &= \mbox{Softmax}\left( \frac{\mathbf{Z}^{(k-1)} \mathbf{W}_{fk} \mathbf{W}_{fq} \mathbf{S}^{(k)}}{\sqrt{d}} \right)
	\end{aligned}
\end{equation}
where $\mathbf{W}_{fk} \in \mathbb{R}^{d_k \times d_c}$, $\mathbf{W}_{fq} \in \mathbb{R}^{d_c \times d_k}$ and $\mathbf{W}_{fv} \in \mathbb{R}^{d_k \times d_k}$.

\textbf{\emph{Graph Information Diffusion}} feeds the supernodes $\widehat{\mathbf{S}}^{(k)}$ back to update node set $\mathbf{Z}^{(k)}$.
With the node embeddings $\mathbf{Z}^{(k-1)}$ and supernode embeddings $\widehat{\mathbf{S}}^{(k)}$,
the backward cross-attention (BCAttn) is defined as,
\begin{equation}
	\begin{aligned}
		\mathbf{Z}^{(k)} &= \mbox{BCAttn} \left( \widetilde{\mathbf{S}}^{(k)}, \mathbf{Z}^{(k-1)} \right) \\
		&= \mathbf{A}^{(k)}_b \mathbf{Z}^{(k-1)} \mathbf{W}_{bv} \\
		\mathbf{A}^{(k)}_b &= \mbox{Softmax} \left( \frac{\widehat{\mathbf{S}}^{(k)} \mathbf{W}_{bq} \mathbf{W}_{bk} \mathbf{Z}^{(k-1)}}{\sqrt{d}} \right)
	\end{aligned}
\end{equation}
where $\mathbf{W}_{bq} \in \mathbb{R}^{d_k \times d_a}$, $\mathbf{W}_{bk} \in \mathbb{R}^{d_a \times d_k}$ and $\mathbf{W}_{bv} \in \mathbb{R}^{d_k \times d_k}$.

The last step is adding the probability transition with $\widetilde{\mathbf{A}}$.
The output of $k$-th order HD-ChebNet (Equation \ref{final_define}) is,
\begin{equation}
	\widehat{\mathbf{Z}}^{(k)} = \sigma \left( \widetilde{\mathbf{A}} \mathbf{Z}^{(k)} \mathbf{W}^{(k)} \right)
\end{equation}

Finally,
the outputs from the prime ChebNet and HD-ChebNets are integrated as the node embeddings,
\begin{equation}
	\mathbf{H} = \mbox{norm} \left( \mathbf{Z}^{(0)} + \sum_{k=1}^{K/2} \widehat{\mathbf{Z}}^{(k)} \right).
\end{equation}

\subsection{Classifier Layer}

\textbf{\emph{Node Classification}}
The node representations $\mathbf{H}$ output from the last graph convolutional layer are straightforward fed into a \emph{Softmax} classifier for node classification.
\begin{equation}
	\widehat{y}_v = \mbox{Softmax}\left( \mbox{MLP} \left( \bm{h}_v \right) \right)
\end{equation}

\textbf{\emph{Graph Classification}}
The representation on the whole graph is constructed via a readout layer on the outputs $\mathbf{H}$,
\begin{equation}
	\centering
	\begin{aligned}
		\bm{h}_{v} &= \sigma \left( f_1 \left(\bm{h}_v \right) \right) \odot \tanh \left( f_2 \left( \bm{h}_v \right) \right) \\ 
		\bm{h}_{g} &= \frac{1}{\vert \mathcal{V} \vert} \sum^{\vert \mathcal{V} \vert}_{v=1} \bm{h}_v + \mbox{Maxpool} \left( \bm{h}_1 \cdots \bm{h}_{\vert \mathcal{V} \vert} \right)
	\end{aligned}
\end{equation}
where $\odot$ denotes the Hadamard product and $f_1 ()$, $f_2 ()$ are two non-linear functions.
 
The graph representation $\bm{h}_g \in \mathbb{R}^d$ is fed into the \emph{Softmax} classifier to predict the graph label.
\begin{equation}
 	\widehat{y}_{g} = \mbox{Softmax}\left( \bm{h}_{g} \right)
\end{equation}

All parameters are optimized by minimizing the cross-entropy function:
\begin{equation}
	\mathcal{L} = - \frac{1}{N} \sum_{n=1}^N y_{n/g} \log (\widehat{y}_{n/g})
\end{equation}

\section{Experiments}

In this section,
we evaluate HDGCN on \emph{transductive} and \emph{inductive} NLP tasks of text classification, aspect-based sentiment classification, natural language inference, and node classification.
In experiment,
each layer of HDGCN is fixed with $K=6$ order Chebyshev approximation and the model stacks $L=1$ layer.
The dimension of input node embeddings is $d=300$ of GlVe or $d=768$ of pre-trained BERT,
and the hyper-parameter $d_k=64$, $d_a=64$.
So the weights $\mathbf{W}^{(0)} \in \mathbb{R}^{d \times 64}$, $\mathbf{W}^{l}_d, \mathbf{W}^{(k)} \in \mathbb{R}^{64 \times 64}$ and $\mathbf{W}_{fk}, \mathbf{W}_{fq}, \mathbf{W}_{bq}, \mathbf{W}_{bk} \in \mathbf{R}^{64 \times 64}$.
The number of super-nodes is set as $M=10$.
Our model is optimized with adaBelief \cite{zhuang2020adabelief} with a learning rate $1e-5$.
The schematics about the HDGCN is shown in Figure \ref{mvcattn_schem}.

To analyze the effectiveness of MVCAttn in avoiding over-smoothing,
we report the results of ablation study - HDGCN-\emph{static} in Table~\ref{tab:results}, \ref{tab:large_scale_text} \ref{transductive}.
The ablation model - HDGCN-\emph{static} is an implementation of Equation \ref{3rdcheb},
in which the node state transition is determined by the static adjacency matrix $\widetilde{\mathbf{A}}^{2k}$.

\subsection{Text Classification}

The first experiment is designed to evaluate the performance of HDGCN on the text graph classification.
Four small-scale text datasets\footnote{https://github.com/yao8839836/text\_gcn} - MR, R8, R52, Ohsumed,
and four large-scale text datasets - AG's News\footnote{http://groups.di.unipi.it/~gulli/AG\_corpus\_of\_news\_articles.html}, SST-1, SST-2\footnote{https://nlp.stanford.edu/sentiment/treebank.html}, Yelp-F\footnote{https://www.yelp.com/dataset} are used in this task.
The graph structures are built on word-word co-occurrences in a sliding window (width=3 and unweighted) on individual documents.
HDGCN is initialized with word embeddings pre-trained by $300$-d GloVe and $768$-d BERT-base on small and large scale datasets respectively.
The baselines include TextCNN, TextRNN, fastText, SWEM, TextGCN, GraphCNN, TextING, minCUT, BERT-base, DRNN, CNN-NSU, CapNets, LK-MTL, TinyBERT, Star-Transformer.

\begin{table}[h]
	\centering
	{\scriptsize
	\begin{tabular}{lccccc}
		\hline
		Model       & MR    & R8    & R52   & Ohsumed \\ \hline
		TextCNN$^\star$ & 77.75 & 95.71 & 87.59 & 58.44 \\
		TextRNN$^\star$ & 77.68 & 96.31 & 90.54 & 49.27 \\
		fastText$^\star$ & 75.14 & 96.13 & 92.81 & 57.70 \\
		SWEM$^\star$ & 76.65 & 95.32 & 92.94 & 63.12 \\
		TextGCN$^\star$ & 76.74 & 97.07 & 93.56 & 68.36 \\
		GraphCNN$^\star$ & -   & 97.80 & 94.60 & 69.40 \\
		minCUT ~\cite{bianchi2019mincut}     & 76.52 & 97.42 & 93.53 & 66.37 \\
		TextING ~\cite{zhang2020every}       & 79.82 & 98.04 & 95.48 & 70.42 \\
		BERT-base \cite{jin2019bert}         & 85.80 & 97.92 & 96.37 & 71.04 \\ \hline
		HDGCN-\emph{static}                  & 79.70 & 98.05 & 95.49 & 70.75 \\
		HDGCN    & \textbf{86.50} & \textbf{98.45} & \textbf{96.57} & \textbf{73.97} \\ \hline
	\end{tabular}
	}
	\caption{Test accuracy (\%) on small-scale English datasets, where the results labeled with $\star$ are cited from \protect\cite{zhang2020every}.}\label{tab:results}
\end{table}
\begin{table}[h]
	\centering
	{\scriptsize
	\begin{tabular}{lcccc}
		\hline
		Model                       & AG        & SST-1 & SST-2 & Yelp-F \\ \hline
		fastText~\citep{joulin2017bag}        & 92.5      & -   & -   & 63.9   \\
		DRNN~\cite{wang2018disconnected}     & 93.6      & 47.3  & 86.4  & 65.3   \\
		CNN-NSU~\cite{li2017initializing}   & -       & 50.8  & 89.4  & -    \\
		CapNets~\cite{yang2018investigating} & 92.6      & -   & 86.8  & -    \\
		LK-MTL~\cite{xiao2018learning}       & -       & 49.7  & 88.5  & -    \\
		BERT~\cite{xiao2018learning}         & 94.5      & 50.1  & 89.3  & 65.8  \\
		TinyBERT~\cite{jiao2020tinybert}       & 94.7      & 51.6  & \textbf{92.6}  & 66.1  \\
		Star-Transformer~\cite{guo2019star}  & -       & 52.9  & -    & -    \\ \hline
		HDGCN-\emph{static}      & 94.0        & 52.1  & 90.8 & 65.2  \\
		HDGCN                    & \textbf{95.5} & \textbf{53.9} & 92.3 & \textbf{69.6} \\ \hline
	\end{tabular}
	}
	\caption{Test accuracies (\%) on large-scale English datasets.}\label{tab:large_scale_text}
\end{table}

Table~\ref{tab:results} shows the test accuracies on four small-scale English datasets,
in which HDGCN ranks top with accuracies 86.50\%, 98.45\%, 96.57\%, 73.97\% respectively.
HDGCN beats the best baselines achieved by TextING (the newest GNN model) and the fine-tuned BERT-base model.
Our ablation model HDGCN-\emph{static} also achieves higher accuracies than the newest GNN models - TextING and minCUT. 
Therefore,
the outperformance of HDGCN verifies that (1) the node probability transition in high-order Chebyshev approximation improves the spectral graph convolution; (2) the MVCAttn mechanism in high-order ChebNet further raises the effectiveness by avoiding the over-smoothing problem.

Table~\ref{tab:large_scale_text} shows the test accuracies of HDGCN and other SOTA models on large-scale English datasets.
HDGCN achieves the best results 95.5\%, 53.9\%, 69.6\% on AG, SST-1, Yelp-F respectively,
and performs a slight gap 0.3\% with the top-1 baseline (TinyBERT) on SST-2.
These results support that HDGCN outperforms the fully-connected graph module in Transformers and corresponding pre-trained models.
Additionally,
these comparisons also demonstrates that the combination of prior graph structures and self-adaptive graph structures in graph convolution is able to improve the multi-hop graph reasoning.

\subsection{Multi-hop Graph Reasoning in Text Graph}

\begin{figure}[h]
	\centering
	\includegraphics[scale=0.055]{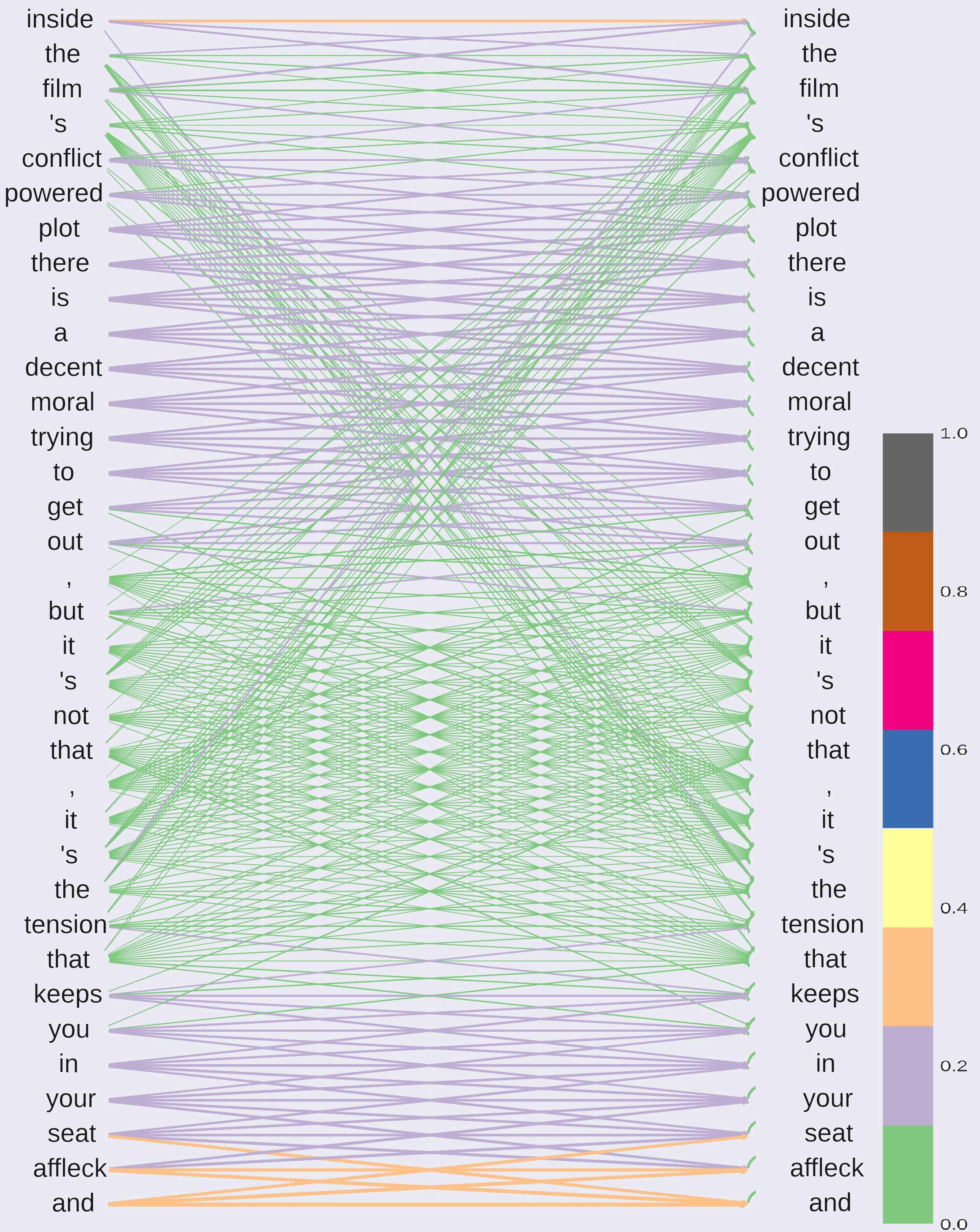}
	\caption{The message aggregation on adjacency matrix $\widetilde{\mathbf{A}}$ with word-word co-occurrence in document.}
	\label{fig:res}
\end{figure}

In the second experiment,
we make a case study on the MR dataset to visualize how the HDGCN improve multi-hop graph reasoning.
Here,
we take the positive comment "\emph{inside the film's conflict powered plot there is a decent moral trying to get out, but it's not that , it's the tension that keeps you in your seat Affleck and Jackson are good sparring partners}" as an example.

First,
the word interactions on prior graph structure $\widetilde{\mathbf{A}}$ (word-word co-occurrence in a sliding window with width=3) is showed in Figure \ref{fig:res}.
We can see that the word mainly interacts with its consecutive neighbors.
It is hard for the vanilla GCN to encode multi-hop and non-consecutive word-word interactions as the example shown in Figure~\ref{mhr_demo}.

\begin{table*}[h]
	\centering
	{\scriptsize
	\begin{tabular}{clccccccccccc}
		\hline
		\multirow{2}{*}{\begin{tabular}[c]{@{}l@{}}Initialized\\ embeddings\end{tabular}} & \multirow{2}{*}{Model} & \multicolumn{2}{c}{TWITTER} & \multicolumn{2}{c}{LAP14} & \multicolumn{2}{c}{REST14} & \multicolumn{2}{c}{REST15} & \multicolumn{2}{c}{REST16} \\ \cline{3-12} 
		& & Acc.         & F1.          & Acc.        & F1.         & Acc.         & F1.         & Acc.         & F1.         & Acc.         & F1          \\ \hline
		\multirow{5}{*}{GloVe} & AOA$^\star$ & 72.30 & 70.20 & 72.62       & 67.52       & 79.97        & 70.42       & 78.17        & 57.02       & 87.50        & 66.21       \\
		& TNet-LF$^\star$ & 72.98  & 71.43 & 74.61       & 70.14       & 80.42        & 71.03       & 78.47         & 59.47       & 89.07          & 70.43 \\
		& ASCNN$^\star$ & 71.05        & 69.45       & 72.62       & 66.72       & 81.73        & 73.10         & 78.47        & 58.90       & 87.39        & 64.56       \\
		& ASGCN-DT$^\star$ & 71.53        & 69.68       & 74.14       & 69.24       & 80.86        & 72.19         & 79.34 & 60.78       & 88.69        & 66.64       \\
		& ASGCN-DG$^\star$ & 72.15        & 70.40        & 75.55       & 71.05       & 80.77        & 72.02       & 79.89 & 61.89 & 88.99 & 67.48       \\ \hline
		\multirow{3}{*}{BERT-base} & AEN-BERT~\cite{song2019attentional}  & - & - & 79.93 & 76.31 & 83.12 & 73.76 & - & - & - & - \\
		& BERT-PT~\cite{xu2019bert}      & - & - & 78.07 & 75.08 & 84.95 & 76.96 & - & - & - & - \\
		& SDGCN-BERT~\cite{zhao2020modeling}      & -   & -   & \textbf{81.35} & \textbf{78.34} & 83.57 & 76.47 & -   & -  & -   & -   \\ \hline
		& HDGCN (GloVe)& \textbf{73.41} & \textbf{71.52} & 76.80 & 73.18 & 80.43        & 70.74       & \textbf{81.18} & \textbf{67.40} & \textbf{89.12} & 70.37 \\
		& HDGCN (BERT-base) & 72.69 & 71.23 & 79.15 & 75.48 & \textbf{85.89} & \textbf{79.33} & \textbf{81.18} & 62.21 & 87.99 & \textbf{71.28} \\ \hline
	\end{tabular}
	}
	\caption{Test accuracy (\%) and macro-F1 score on aspect-based sentiment classification. The results labeled with $^\star$ are cited from \protect\cite{zhao2020modeling}.}
	\label{aspect_analysis}
\end{table*}

\begin{figure}
	\centering
	\includegraphics[scale=0.05]{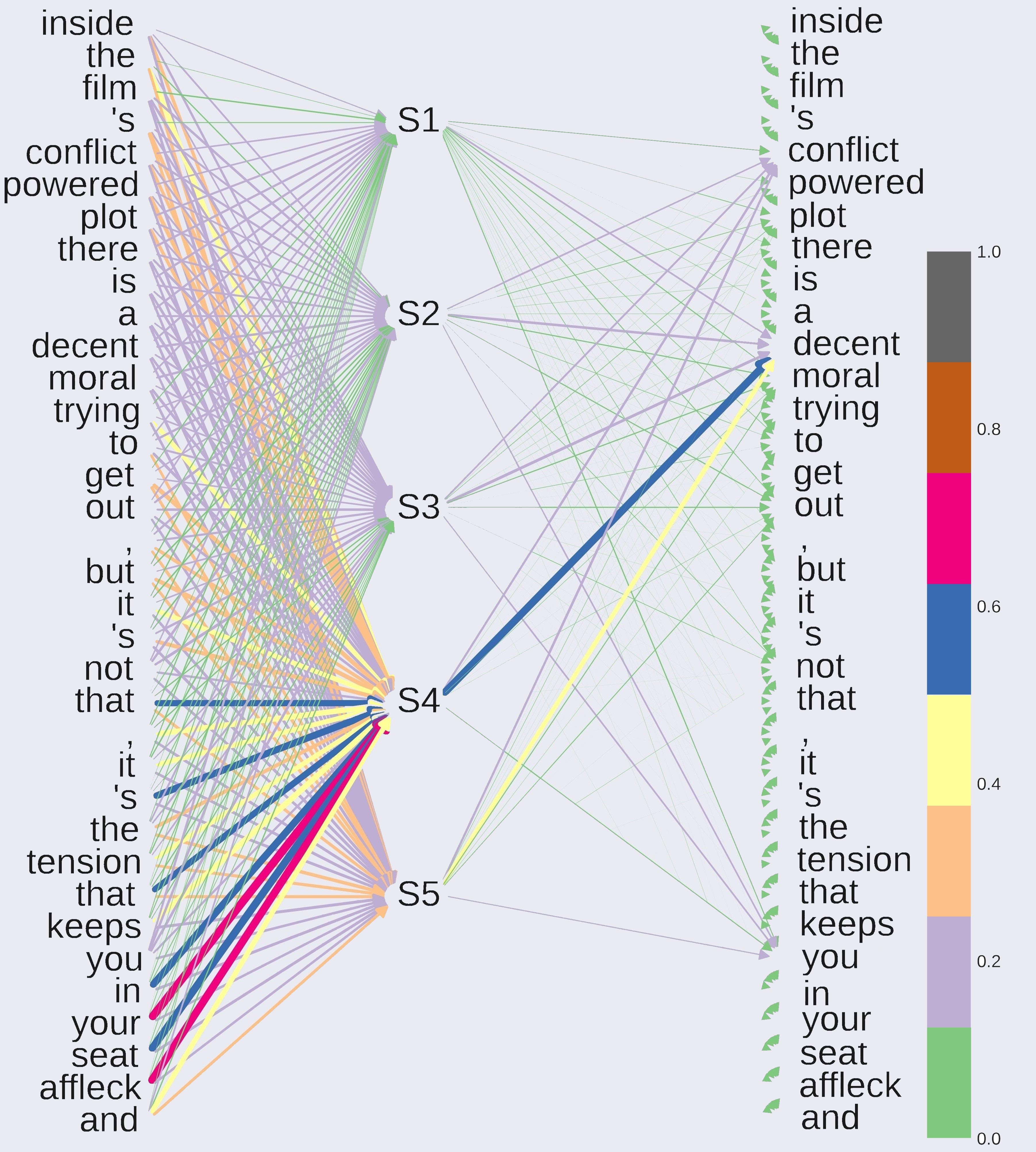}
	\caption{The word interactions in MVCAttn ($\mathbf{A}^{(1)}_f$ $\times$ $\mathbf{A}^{(1)}_b$) of the 1st HD-ChebNet, where $S1 \sim S5$ represent the supernodes.}
	\label{layer1bipartite}
\end{figure}

Figure \ref{layer1bipartite} shows the node interactions from node embeddings $\mathbf{Z}^{(0)}$ to supernodes $\widehat{\mathbf{S}}^{(1)}$ and the graph diffusion from $\widehat{\mathbf{S}}^{(1)}$ to node embeddings $\mathbf{Z}^{(1)}$.
In which,
the supernode $S4$ puts greater attention on the segment - \emph{it's the tension that keeps you in your seat}.
This segment determines its positive polarity significantly.
The other supernodes $S1$, $S2$, $S3$, $S5$ just aggregate messages from the global context evenly.
Next,
the messages aggregated in supernodes $S1 \sim S5$ are mainly diffused to four tokens - \emph{conflict}, \emph{decent}, \emph{moral}, \emph{you}.
That verifies the self-adaptively learned graph structure $\mathbf{A}^{(1)}_f \times \mathbf{A}^{(1)}_b$ by the MVCAttn improves the multi-hop graph reasoning on nodes - \emph{conflict, decent, moral, you}.
From the perspective of semantics,
these four words determine the positive sentiment in this comment significantly.

\begin{figure}
	\centering
	\includegraphics[scale=0.045]{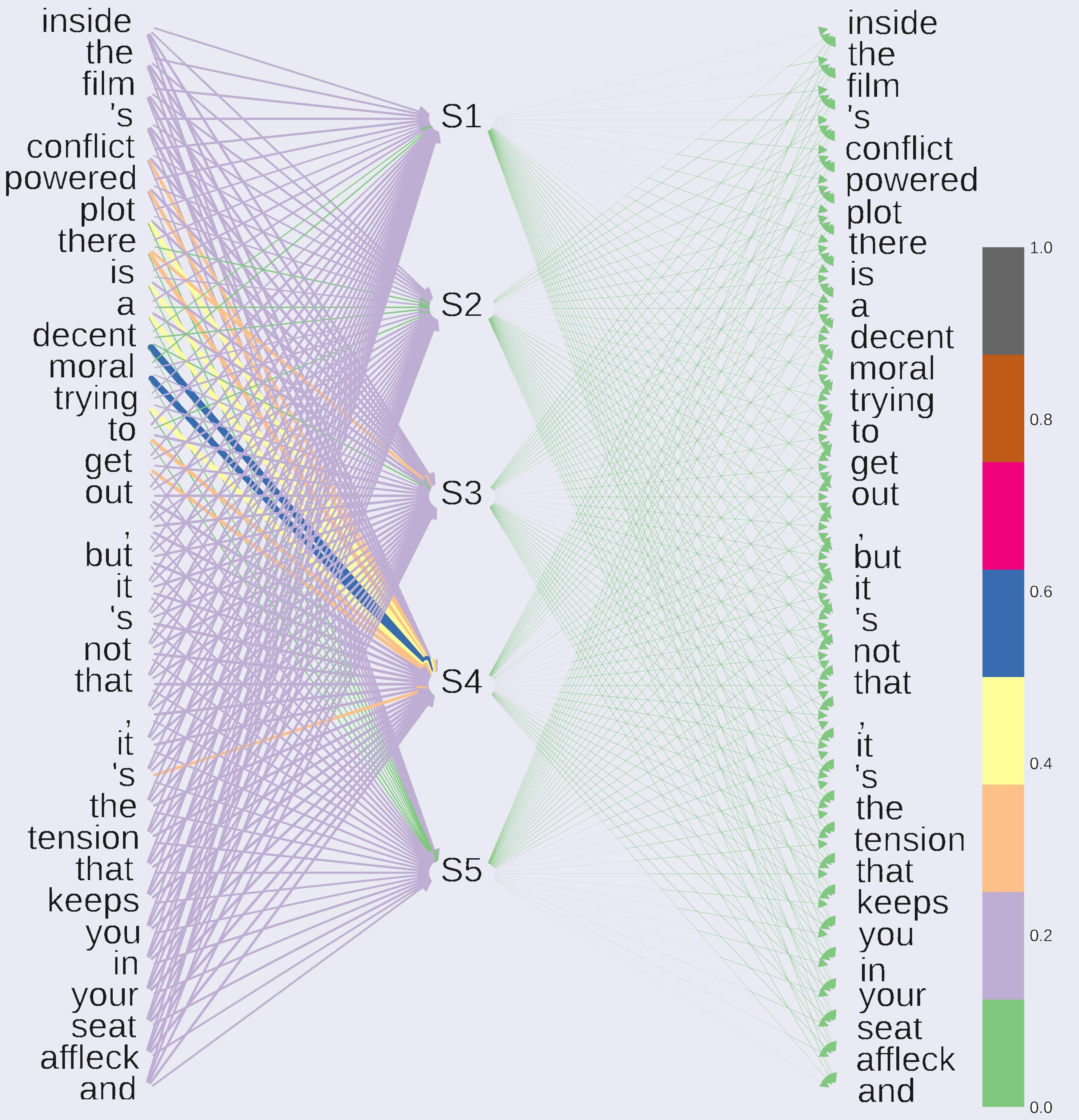}
	\caption{The word interactions in MVCAttn $\mathbf{A}^{(2)}_f$ $\times$ $\mathbf{A}^{(2)}_b$ of the 2nd HD-ChebNet, where $S1 \sim S5$ represent the supernodes.}
	\label{layer2bipartite}
\end{figure}

Figure \ref{layer2bipartite} shows the message aggregation from node embeddings $\mathbf{Z}^{(1)}$ to supernodes $\widehat{\mathbf{S}}^{(2)}$ and the message diffusion from $\widehat{\mathbf{S}}^{(2)}$ to node embeddings $\mathbf{Z}^{(2)}$.
We can see that the supernode $S4$ puts greater attention on another segment - \emph{there is a decent moral young to get out},
which also contributes to the sentiment polarity.
Then messages aggregated to supernodes $S1 \sim S5$ are diffused to all words evenly.
The backward interactions from supernodes $S1 \sim S5$ to all graph nodes do not have visible differences.
These results demonstrate that the multi-hop graph reasoning in HDGCN just needs one graph convolutional layer to attain the stationary state.

\subsection{Aspect-based Sentiment Classification}

\begin{figure}[h]
	\centering
	\includegraphics[scale=0.035]{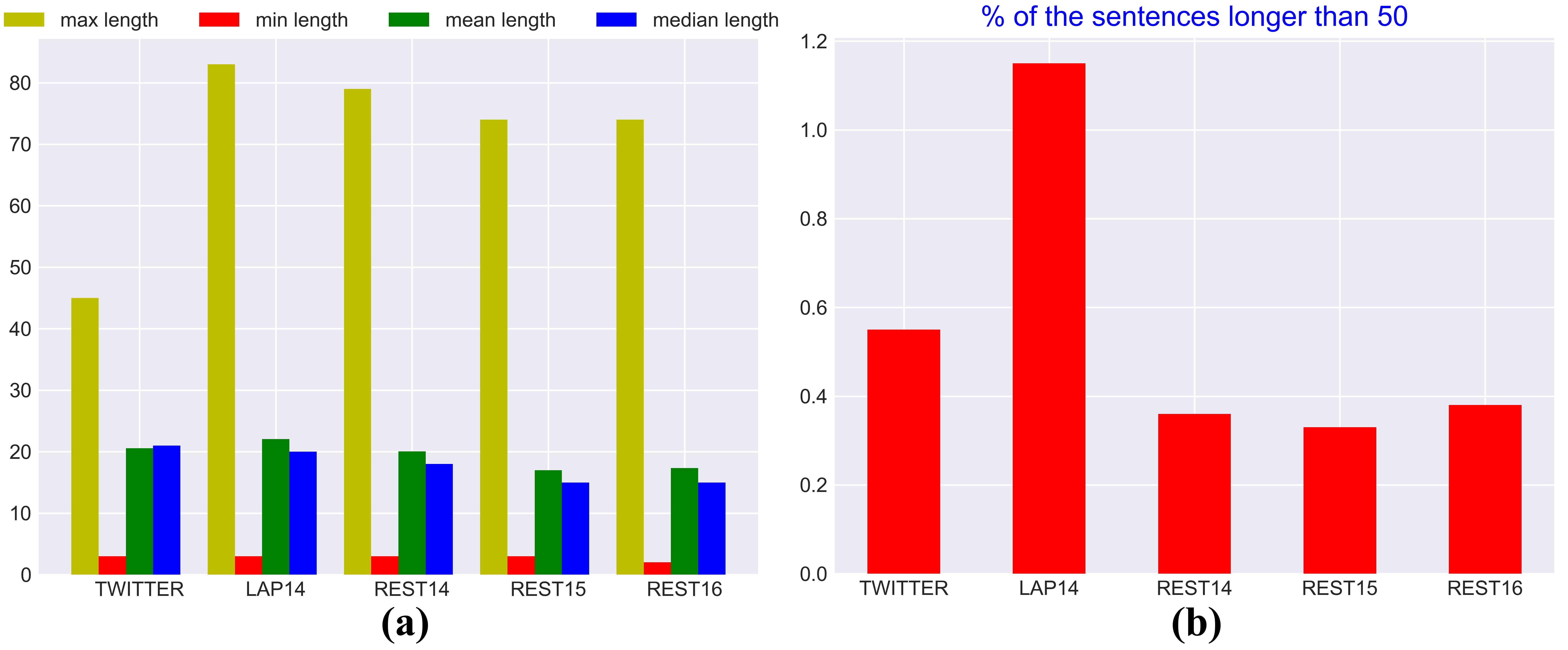}
	\caption{\textbf{(a):} The statistics of aspect-based sentiment classification datasets. \textbf{(b):} The percentages of sentences with length $\geq 50$ in 5 datasets.}
	\label{dataset_statistics}
\end{figure}

The third experiment evaluates HDGCN's performance on the task of aspect-based sentiment classification.
This task aims to identify whether the sentiment polarities of aspect are explicitly given in sentences~\cite{zhao2020modeling}.
The datasets used in this task include TWITTER, LAP14, REST14, REST15, REST16 ~\cite{zhao2020modeling}.
The details about the statistics on these datasets are shown in Figure \ref{dataset_statistics}.
The SOTA comparison models include AOA, TNet-LF, ASCNN, ASGCN-DT, ASGCN-DG, AEN-BERT, BERT-PT, SDGCN-BERT.

Each sample in this task includes a sentence pair, an aspect, and a label.
The sentence pair and the aspect are concatenated into one long sentence,
and the text graph is preprocessed with the dependency tree on this sentence.
HDGCN is tested twice with word embeddings initialized by pre-trained $300$-d GloVe and $768$-d BERT-base respectively.

Table~\ref{aspect_analysis} shows the test accuracies and micro-F1 scores on 5 datasets,
where HDGCN achieves new state-of-the-art results on TWITTER, REST14, REST15, REST16,
and a top-3 result on the LAP14.
As shown in Figure \ref{dataset_statistics} that the LAP14 has the maximum percentage of long sentences among all datasets.
A shallow network in HDGCN does not outperform the SOTA result on the LAP14.
Additionally,
compared with the newest ASGCN and attention-based AOA,
HDGCN achieves the best results on TWITTER, LAP14, REST15, REST16 (Acc) and performs very close with the highest accuracy on REST14 and macro-F1 score on REST16.
Above comparison supports that the matching between aspect and sentence pair in HDGCN is more accurate than the newest GNN and attention-based models,
which verifies that the multi-hop graph reasoning is improved in HDGCN. 

\subsection{Natural Language Inference}

The fourth experiment evaluates HDGCN's performance on the Stanford natural language inference (SNLI) task~\cite{bowman2015large}.
This task aims to predict the semantic relationship is \underline{\emph{entailment}} or \underline{\emph{contradiction}} or \underline{\emph{neutral}} between a premise sentence and a hypothesis sentence.
All the comparison methods include fine-tuned BERT-base, MT-DNN~\cite{liu2020microsoft}, SMART~\cite{jiang2020smart}, and CA-MTL~\cite{pilault2020conditionally}.

In this task,
the premise and hypothesis sentences are concatenated and constructed into a long sentence.
Which is preprocessed to a text graph with the dependency tree.
The word embeddings in HDGCN were initialized from the pre-trained $768$-d BERT-base.

\begin{table}[h]
\centering
{\scriptsize
\begin{tabular}{lcccc}
\hline
\multicolumn{1}{l}{\multirow{2}{*}{Model}} & \multicolumn{1}{l}{\multirow{2}{*}{{\begin{tabular}[c]{@{}c@{}}Total\\ parameters\end{tabular}}}} & \multicolumn{3}{c}{\% data used} \\ \cline{3-5} 
\multicolumn{1}{c}{}                   &  &0.1\% & 1.0\% & 10\% \\ \hline
BERT-base~\cite{devlin2019bert}        & $1.0 \times$ & 52.5 & 78.1 & 86.7 \\
MT-DNN~\cite{liu2020microsoft}         & - & 81.9 & \textbf{88.3} & 91.1 \\
SMART~\cite{jiang2020smart}            & - & 82.7 & 86.0 & 88.7 \\
CA-MTL~\cite{pilault2020conditionally} & $1.12 \times$ & \textbf{82.8} & 86.2 & 88.0 \\ \hline
HDGCN                                  & $1.02 \times$ & 80.3 & 85.6 & \textbf{92.3} \\ \hline
\end{tabular}
}\caption{Test accuracy (\%) on SNLI, where the total parameters take the BERT-base as base.}\label{snli}
\end{table}

All test accuracies are shown in Table~\ref{snli},
where HDGCN achieves the new state-of-the-art results on the $10\%$ data.
As the MT-DNN, SMART and CA-MTL are all fine-tuned on multi-task learning,
they perform better than HDGCN in low resource regimes ($0.1\%$ and $1.0\%$ of the data). 
HDGCN just uses $0.02 \times$ more parameters than the BERT-base,
and it outperforms the later model on all scales of data.
These results verify that the combination of prior graph structure and self-adaptive graph structure in HDGCN performs comparably with the fully-adaptive graph structures in Transformers and BERT-based multi-task learning models.

\subsection{Graph Node Classification}

The fifth experiment evaluates the effectiveness of HDGCN on the node classification task.
We use three standard citation network benchmark datasets - Cora, Citeseer, and Pubmed, to compare the test accuracies on \emph{transductive} node classification.
In the three datasets,
the nodes represent the documents and edges (undirected) represent citations.
The node features correspond to elements of a bag-of-words representation of a document~\cite{velivckovic2018graph}.
We also use the PPI dataset to compare the results on \emph{inductive} node classification,
which consists of graphs corresponding to different human tissues.
The baselines for comparison include GCN, GAT, Graph-Bert, GraphNAS, LoopyNet, HGCN, GRACE, GCNII.
The results of our evaluation are recorded in Table~\ref{transductive}.

\begin{table}[h]
	\centering
	{\tiny
\begin{tabular}{lccc|c}
\hline
           & \multicolumn{3}{c|}{\emph{\textbf{Transductive}}} & \emph{\textbf{Inductive}}  \\
           & \multicolumn{3}{c|}{(ACC, \%)}    & (micro-F1) \\ \hline
Model      & Cora    & Citeseer    & Pubmed    & PPI        \\ \hline
GCN~\cite{kipf2016semi}          & 85.8    & 73.7        & 88.1      & 69.7       \\
GAT~\cite{velivckovic2018graph}  & 86.4    & 74.3        & 87.6      & 97.3       \\
Graph-Bert~\cite{zhang2020graph} & 84.3    & 71.2        & 79.3      & -          \\
GraphNAS~\cite{gao2019graphnas}  & 84.2    & 73.1        & 79.6      & 98.6          \\
LoopyNet~\cite{zhang2019gresnet} & 83.9    & 73.7        & 83.0      & -          \\
HGCN~\cite{chami2019hyperbolic}  & 79.9   & -           & 80.3    & 74.6       \\
GRACE~\cite{zhu2020deep}      & 83.3    & 72.1  & 86.7                      & 96.9          \\
GCNII~\cite{chen2020simple}   & 86.4  & 76.5  & 85.6       & \textbf{99.5}                \\ \hline
HDGCN-\emph{static}           & 84.2 & 73.2 & 90.3 & 50.4 \\
HDGCN   & \textbf{88.6}    & \textbf{77.0}      & \textbf{91.0}      & \textbf{99.5}      \\ \hline
\end{tabular}
	\caption{Test accuracy ($\%$) on Cora, Citeseer, Pubmed and micro-F1 score ($\%$) on PPI.}\label{transductive}
	}
\end{table}

HDGCN achieves the new state-of-the-art results on Cora, Citeseer and Pubmed,
and performs equally best with the newest GCNII on PPI.
Our ablation model, HDGCN-\emph{static}, also achieves close results with the newest GNNs on Cora, Citeseer, Pubmed,
but it performs poorly on PPI.
Which verifies that the high-order Chebyshev approximation of spectral graph convolution has more serious over-smoothing problem in \emph{inductive} node classification than \emph{transductive} node classification.
All comparisons in this experiment demonstrate the effectiveness of MVCAttn to avoid the over-smoothing problem.

\section{Conclusions}

This study proposes a multi-hop graph convolutional network on high-order dynamic Chebyshev approximation (HDGCN) for text reasoning.
To improve the multi-hop graph reasoning,
each convolutional layer in HDGCN fuses low-pass signals (direct dependencies saved in fixed graph structures) and high-pass signals (multi-hop dependencies adaptively learned by MVCAttn) simultaneously.
We also firstly propose the multi-votes based cross-attention (MVCAttn) mechanism to alleviate the over-smoothing in high-order Chebyshev approximation,
and it just costs the linear computation complexity.
Our experimental results demonstrate that HDGCN outperforms compared SOTA models on multiple transductive and inductive NLP tasks.

\section*{Acknowledgments}

This work is supported by Natural Science Foundation of China (Grant No.61872113, 62006061),
Strategic Emerging Industry Development Special Funds of Shenzhen (Grant No.XMHT20190108009),
the Tencent Group Science and Technology Planning Project of Shenzhen (Grant No.JCYJ20190806112210067) and Shenzhen Foundational Research Funding (Grant No.JCYJ20200109113403826).

\bibliographystyle{acl_natbib}
\bibliography{anthology,acl2021}

\onecolumn
\appendix
\section{Appendices}
Here,
we give the completed proof about our high-order Chebyshev approximation on the spectral graph convolution.
We exhibit how to deduce the spectral graph convolution to 4th-order Chebyshev polynomials as follows.

\begin{equation*}
	\bm{g}_{\boldsymbol{\theta}} \star \bm{x} = \mathbf{U} \bm{g}_{\boldsymbol{\theta}} \mathbf{U}^T \bm{x}
\end{equation*}
where $\bm{g}_{\boldsymbol{\theta}}=\bm{g}_{\boldsymbol{\theta}}(\widetilde{\boldsymbol{\Lambda}})$ is the graph filter defined in spectral domain.

\begin{equation*}
	\begin{aligned}
		\bm{g}_{\boldsymbol{\theta }} &\approx \boldsymbol{\theta }_0+\boldsymbol{\theta }_1\widetilde{\mathbf{\Lambda }}+\boldsymbol{\theta }_2\left( 2\widetilde{\mathbf{\Lambda }}^2-1 \right) +\boldsymbol{\theta }_3\left( 4\widetilde{\boldsymbol{\varLambda }}^3-3\widetilde{\mathbf{\Lambda }} \right) \\
		&=\boldsymbol{\theta }_0+\boldsymbol{\theta }_1\widetilde{\mathbf{\Lambda }}+2\boldsymbol{\theta }_2\widetilde{\mathbf{\Lambda }}^2-\boldsymbol{\theta }_2+4\boldsymbol{\theta }_3\widetilde{\mathbf{\Lambda }}^3-3\boldsymbol{\theta }_3\widetilde{\mathbf{\Lambda }}
	\end{aligned}
\end{equation*}

So,

\begin{equation*}
{\small
	\begin{aligned}
		\mathbf{U}\boldsymbol{g}_{\boldsymbol{\theta }}\mathbf{U}^{\mathrm{T}}\boldsymbol{x} &\approx \boldsymbol{\theta }_0\boldsymbol{x}+\boldsymbol{\theta }_1\mathbf{U}\widetilde{\mathbf{\Lambda }}\mathbf{U}^{\mathrm{T}}\boldsymbol{x}+2\boldsymbol{\theta }_2\mathbf{U}\widetilde{\mathbf{\Lambda }}^2\mathbf{U}^{\mathrm{T}}\boldsymbol{x} - \boldsymbol{\theta }_2\boldsymbol{x}+4\boldsymbol{\theta }_3\mathbf{U}\widetilde{\mathbf{\Lambda }}^3\mathbf{U}^{\mathrm{T}}\boldsymbol{x}-3\boldsymbol{\theta }_3\mathbf{U}\widetilde{\mathbf{\Lambda }}\mathbf{U}^{\mathrm{T}}\boldsymbol{x} \\
		&=\boldsymbol{\theta }_0\boldsymbol{x}+\boldsymbol{\theta }_1\mathbf{U}\widetilde{\mathbf{\Lambda }}\mathbf{U}^{\mathrm{T}}\boldsymbol{x}+2\boldsymbol{\theta }_2\mathbf{U}\widetilde{\mathbf{\Lambda }}\mathbf{U}^{\mathrm{T}}\mathbf{U}\widetilde{\mathbf{\Lambda }}\mathbf{U}^{\mathrm{T}}\boldsymbol{x} - \boldsymbol{\theta }_2\boldsymbol{x}+4\boldsymbol{\theta }_3\mathbf{U}\widetilde{\mathbf{\Lambda }}\mathbf{U}^{\mathrm{T}}\mathbf{U}\widetilde{\mathbf{\Lambda }}\mathbf{U}^{\mathrm{T}}\mathbf{U}\widetilde{\mathbf{\Lambda }}\mathbf{U}^{\mathrm{T}}\boldsymbol{x}-3\boldsymbol{\theta }_3\mathbf{U}\widetilde{\mathbf{\Lambda }}\mathbf{U}^{\mathrm{T}}\boldsymbol{x} \\
		&=\mathbf{U}\widetilde{\mathbf{\Lambda }}\mathbf{U}^{\mathrm{T}}\mathbf{U}\widetilde{\mathbf{\Lambda }}\mathbf{U}^{\mathrm{T}} ( \frac{\boldsymbol{\theta }_0}{\mathbf{U}\widetilde{\mathbf{\Lambda }}\mathbf{U}^{\mathrm{T}}\mathbf{U}\widetilde{\mathbf{\Lambda }}\mathbf{U}^{\mathrm{T}}}+\frac{\boldsymbol{\theta }_1}{\mathbf{U}\widetilde{\mathbf{\Lambda }}\mathbf{U}^{\mathrm{T}}} +2\boldsymbol{\theta }_2-\frac{\boldsymbol{\theta }_2}{\mathbf{U}\widetilde{\mathbf{\Lambda }}\mathbf{U}^{\mathrm{T}}\mathbf{U}\widetilde{\mathbf{\Lambda }}\mathbf{U}^{\mathrm{T}}}+4\boldsymbol{\theta }_3\mathbf{U}\widetilde{\mathbf{\Lambda }}\mathbf{U}^{\mathrm{T}}-3\frac{\boldsymbol{\theta }_3}{\mathbf{U}\widetilde{\mathbf{\Lambda }}\mathbf{U}^{\mathrm{T}}} ) \boldsymbol{x} \\
		&=\mathbf{U}\widetilde{\mathbf{\Lambda }}\mathbf{U}^{\mathrm{T}}\mathbf{U}\widetilde{\mathbf{\Lambda }}\mathbf{U}^{\mathrm{T}}\left( \frac{\boldsymbol{\theta }_0-\boldsymbol{\theta }_2}{\mathbf{U}\widetilde{\mathbf{\Lambda }}\mathbf{U}^{\mathrm{T}}\mathbf{U}\widetilde{\mathbf{\Lambda }}\mathbf{U}^{\mathrm{T}}}+\frac{\boldsymbol{\theta }_1-3\boldsymbol{\theta }_3}{\mathbf{U}\widetilde{\mathbf{\Lambda }}\mathbf{U}^{\mathrm{T}}} \right) \boldsymbol{x} + \mathbf{U}\widetilde{\mathbf{\Lambda }}\mathbf{U}^{\mathrm{T}}\mathbf{U}\widetilde{\mathbf{\Lambda }}\mathbf{U}^{\mathrm{T}}\left( 2\boldsymbol{\theta }_2+4\boldsymbol{\theta }_3\mathbf{U}\widetilde{\mathbf{\Lambda }}\mathbf{U}^{\mathrm{T}} \right) \boldsymbol{x} \\
		&=\left( \left( \boldsymbol{\theta }_0-\boldsymbol{\theta }_2 \right) +\left( \boldsymbol{\theta }_1-3\boldsymbol{\theta }_3 \right) \mathbf{U}\widetilde{\mathbf{\Lambda }}\mathbf{U}^{\mathrm{T}} \right) \boldsymbol{x} +\mathbf{U}\widetilde{\mathbf{\Lambda }}\mathbf{U}^{\mathrm{T}}\mathbf{U}\widetilde{\mathbf{\Lambda }}\mathbf{U}^{\mathrm{T}}\left( 2\boldsymbol{\theta }_2+4\boldsymbol{\theta }_3\mathbf{U}\widetilde{\mathbf{\Lambda }}\mathbf{U}^{\mathrm{T}} \right) \boldsymbol{x}
	\end{aligned}}
\end{equation*}

Let assume $\boldsymbol{\theta}^{(0)}=\boldsymbol{\theta}_0 - \boldsymbol{\theta}_2 = - \boldsymbol{\theta}_1 + 3 \boldsymbol{\theta}_3$,
$\boldsymbol{\theta}^{(1)}=2\boldsymbol{\theta}_1 = -4 \boldsymbol{\theta}_3$,
\begin{equation*}
	\begin{aligned}
		\mathbf{U}\bm{g}_{\boldsymbol{\theta }}\mathbf{U}^{\mathrm{T}}\bm{x} &\approx \boldsymbol{\theta }^{\left( 0 \right)}\left( \mathbf{I}-\mathbf{U}\widetilde{\mathbf{\Lambda }}\mathbf{U}^{\mathrm{T}} \right) \boldsymbol{x} +\mathbf{U}\widetilde{\mathbf{\Lambda }}\mathbf{U}^{\mathrm{T}}\mathbf{U}\widetilde{\mathbf{\Lambda }}\mathbf{U}^{\mathrm{T}}\boldsymbol{\theta }^{\left( 1 \right)}\left( \mathbf{I}-\mathbf{U}\widetilde{\mathbf{\Lambda }}\mathbf{U}^{\mathrm{T}} \right) \bm{x} \\
		&\approx \boldsymbol{\theta }^{\left( 0 \right)}\widetilde{\mathbf{A}}\bm{x}+ \widetilde{\mathbf{A}}^2 \boldsymbol{\theta }^{\left( 1 \right)}\widetilde{\mathbf{A}}\bm{x}
	\end{aligned}
\end{equation*}

To avoid the over-smoothing problem in the node state transition $\widetilde{\mathbf{A}}^2$,
the graph structure $\widetilde{\mathbf{A}}$ is approximated by the dynamic adjacency matrix $\mathbf{A}_d$ self-adaptively learned with attention mechanism.
This way have the hidden pairwise interactions to improve the multi-hop graph reasoning in high-order Chebyshev polynomials.
Therefore,
we define the layer-wise propagation of multi-hop graph convolutional network as follows.
\begin{equation*}
	\begin{aligned}
		\mathbf{H} &\approx \sum^{K/2}_{k=0} \mathbf{Z}^{(k)}, \\
		\mathbf{Z}^{(0)} &= \underset{\mbox{Prime ChebNet}}{\underbrace{\sigma \left( \widetilde{\mathbf{A}} \mathbf{X} \mathbf{W}^{(0)} \right)}}, \\
		\mathbf{Z}^{(k)} &= \underset{\mbox{HD-ChebNet}}{\underbrace{\sigma \left( \widetilde{\mathbf{A}} \left( \mathbf{A}_{d}^{(k)} \mathbf{Z}^{(k-1 )}\mathbf{W}^{(k)}_d \right) \mathbf{W}^{(k)} \right)}}
	\end{aligned}
	\label{final_define}
\end{equation*}
where $\mathbf{X}$ is the input features,
and we introduce two nonlinear filterings $\mathbf{W}^{(k)}$ and $\mathbf{W}^{(k)}_d$ on node signals.

\end{document}